\documentclass{article}
\usepackage{spconf,amsmath,graphicx}
\usepackage{algorithm,algorithmic}
\usepackage{url}
\usepackage{subfigure}
\title{Self-augmented multi-modal feature embedding}
\name{Shinnosuke Matsuo, Seiichi Uchida, Brian Kenji Iwana\vspace{-3mm}\thanks{This work was partially supported by JSPS KAKENHI Grant Number JP17H06100.
}}
\address{Kyushu University, Fukuoka, Japan}
\begin{document}
\ninept
\maketitle
\begin{abstract}
Oftentimes, patterns can be represented through different modalities. For example, leaf data can be in the form of images or contours. Handwritten characters can also be either online or offline. To exploit this fact, we propose the use of self-augmentation and combine it with multi-modal feature embedding. In order to take advantage of the complementary information from the different modalities, the self-augmented multi-modal feature embedding employs a shared feature space. Through experimental results on classification with online handwriting and leaf images, we demonstrate that the proposed method can create effective embeddings.
\end{abstract}
\section{Introduction}
\label{sec:intro}

We can often find patterns, such as signals, that have an inherent multi-modality. 
For example, a handwriting trajectory signal, which is generated as a time series (a sequence of the pen-tip coordinates), is, at the same time, observed as a stroke image on a two-dimensional plane. 
In another example, a leaf image is an image sample and, at the same time, can be converted as a sequence of continuous contour point locations. \par
\begin{figure}
    \centering
    \includegraphics[width=0.8\linewidth]{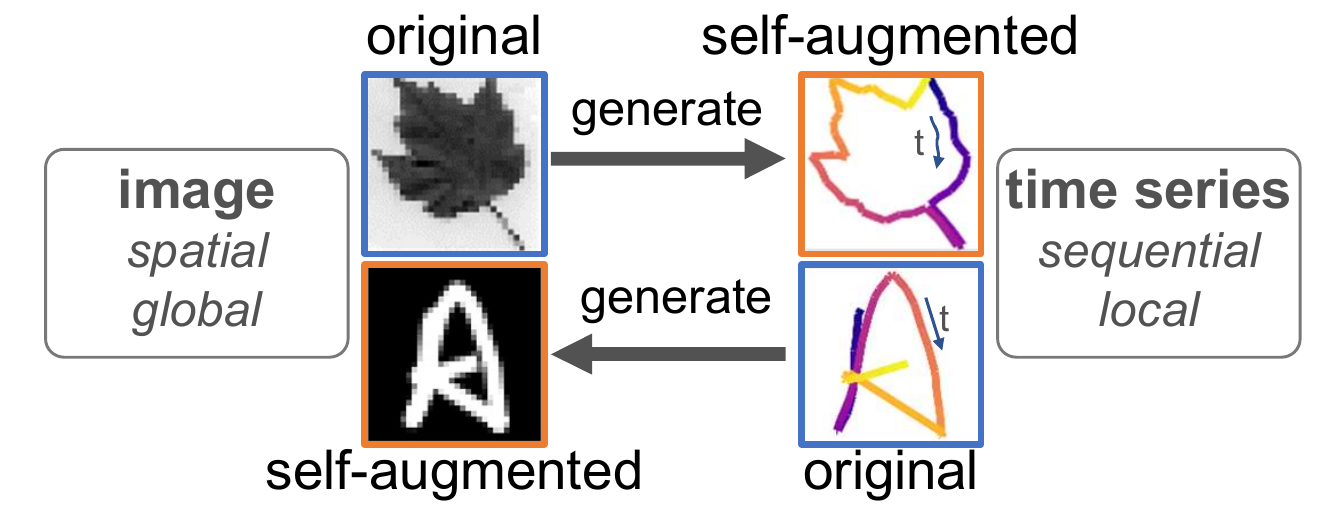}\\[-3mm]
    \caption{The idea of self-augmented multi-modality.}
    \label{fig:concept}
\end{figure}

Accordingly, as illustrated in Fig.~\ref{fig:concept}, it is possible to generate a {\em self-augmented modality} $x_\mathrm{aug}$ from the original modality $x_\mathrm{org}$ and use $x_\mathrm{aug}$ as complementary information.
Through the use of the self-augmented modality, each pattern now has two corresponding modalities. 
For example, a handwriting trajectory signal has a time series input $x_\mathrm{org}$ which maps the temporal changes of the pen-tip positions over time and a self-augmented image modality which represents the spatial relationship between data points without respect to time.\par

\begin{figure}[t]
    \centering
    \includegraphics[width=1\linewidth]{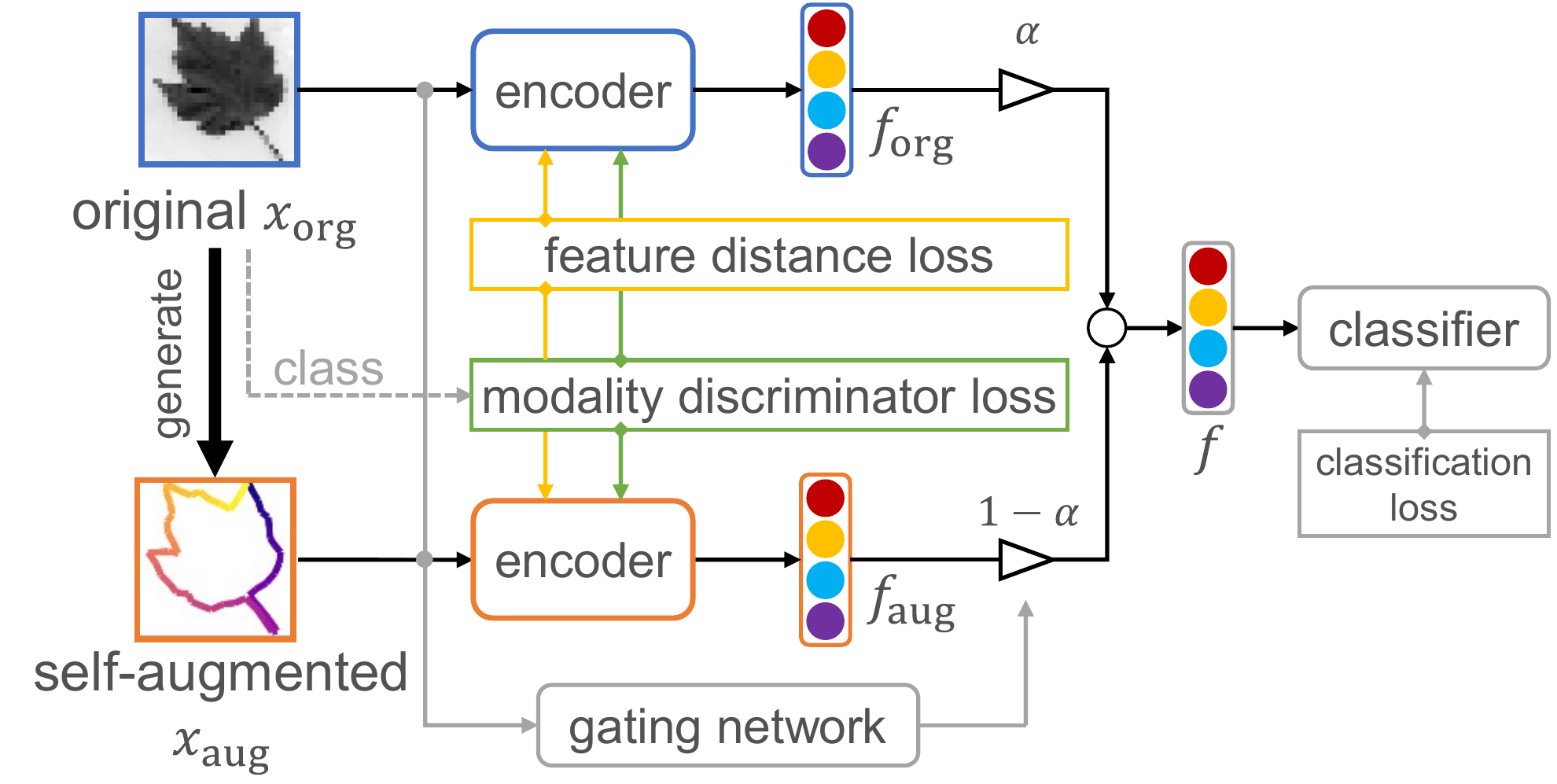} \\[-3mm]
    \caption{The overall structure of the self-augmented multi-modal feature embedding with its application to pattern classification.}
    \label{fig:overall}\vspace{-2mm}
\end{figure}

We propose a method of embedding the patterns in a shared feature space $\mathcal{F}$ between the original and self-augmented modalities. As shown in Fig.~\ref{fig:overall}, the shared feature space uses information from both modalities by developing the following two consistencies between the embeddings:
\begin{itemize}\setlength{\itemsep}{0pt}
\item {\em Hard consistency}. 
Since they originally came from the same pattern, the feature embedding $f_\mathrm{org} \in \mathcal{F}$ of the original pattern $x_\mathrm{org}$ should be the same or similar to the feature embedding $f_\mathrm{aug}\in \mathcal{F}$ of the self-augmented pattern $x_\mathrm{aug}$, or $f_\mathrm{org}\sim f_\mathrm{aug}$.
To accomplish this, we use a distance-based loss that unifies $f_\mathrm{org}$ and $f_\mathrm{aug}$.
\item {\em Soft consistency}. 
Since $f_\mathrm{org}$ and $f_\mathrm{aug}$ are extracted from different modalities, they should keep some of the useful and complementary characteristics of their modalities. 
For these softly-consistent embeddings, an adversarial modality discriminator is used to gently unify the features under class conditions. 
\end{itemize}
These two consistencies provide the embeddings the ability to represent the data irrespective of which modality it came from while still maintaining complementary information.\par
The two embeddings $f_\mathrm{org}$ and  $f_\mathrm{aug}$ are then combined using a gating mechanism trained for some application, such as classification.
Specifically, as shown in Fig.~\ref{fig:overall}, the embeddings from each modality are combined in a feature vector $f\in \mathcal{F}$ based on the gate. 
The gate is controlled using a weight $\alpha$ learned by a simultaneously trained gating network.
This operation is possible because the aforementioned consistencies encourage an element-wise correspondence in $f_\mathrm{org}$ and $f_\mathrm{aug}$.
Our experimental results will show that this operation outperforms a typical network fusion through concatenation in classification tasks.\par

The main contributions of this paper are summarized as follows:
\begin{itemize}\setlength{\itemsep}{0pt}
    \item A method of representing data using self-augmented multi-modal feature embedding using a shared feature space is proposed. The advantage of this embedding is that it is influenced by both modalities and more discriminative.
    \item For classification, a gating network is proposed. Since the proposed embeddings have corresponding features between the modalities, the gating network can simply control the information provided to the classifier. 
    \item Quantitative and qualitative analysis is performed on two datasets with different original modalities, i.e., a time series signal (online handwritten digits and characters) and an image-based pattern (leaf images).  
\end{itemize}

\section{Related Work}
\label{sec:related}
There have been a wide variety of neural network-based multi-modal methods and applications. 
These methods can be roughly categorized into three groups, methods that use network fusion (concatenation)~\cite{duong2017multimodal,ortega2019multimodal,lucieri2020benchmarking}, methods that use gating~\cite{arevalo2017gated}, and methods that use cross-modal training~\cite{park2016image}.
For time series, it is especially common to combine text with images in document recognition~\cite{lucieri2020benchmarking,arevalo2017gated},  natural scene image recognition~\cite{duong2017multimodal}, and cross-modal retrieval~\cite{Wei_2020,Wang_2017adversarial}. 
Combining audio with video is another common use for multi-modal networks~\cite{ortega2019multimodal,Wimmer2008,brady2016multi}. %

Another application of multi-modal networks is to create embeddings for cross-modal generation. 
Some approaches to create shared embeddings include weight-sharing constraints~\cite{Peng_2019,Wang_2019}, multi-view fusion~\cite{Huang_2018}, and cross-modal training~\cite{Spurr_2018,sumi2019modality}.
Similar to the proposed method, adversarial learning has also been used but for generation~\cite{Peng_2019,x-gacmn}. 

The proposed method uses a self-augmented modality, and since the multi-modality is generated from a self-augmented modality, the feature spaces can be shared. We also propose using two consistencies (hard consistency and soft consistency).

\section{Self-Augmented Multi-Modal Feature Embedding}
\label{sec:proposed}
In order to complement the information stored in original pattern $x_\mathrm{org}$, a corresponding cross-modality self-augmented pattern $x_\mathrm{aug}$ is generated, as shown in Fig.~\ref{fig:concept}. For example, tracing the visual contour of an image pattern will generate a self-augmented time-series pattern.  
Since both modalities represent the same pattern, we embed the patterns into a shared feature space $\mathcal{F}$. 
This ensures that the final embeddings are influenced by both modalities. \par

To create the shared feature space $\mathcal{F}$, two parallel encoders, one from each modality, are simultaneously trained but with an extra requirement that the encodings, or embeddings, should be consistent irrespective of the modality. As noted Section~\ref{sec:intro}, hard and soft consistencies between embeddings $f_\mathrm{org}$ and $f_\mathrm{aug}$ are required for the encoders.

\begin{figure}[t]
    \centering
    \includegraphics[width=1\linewidth]{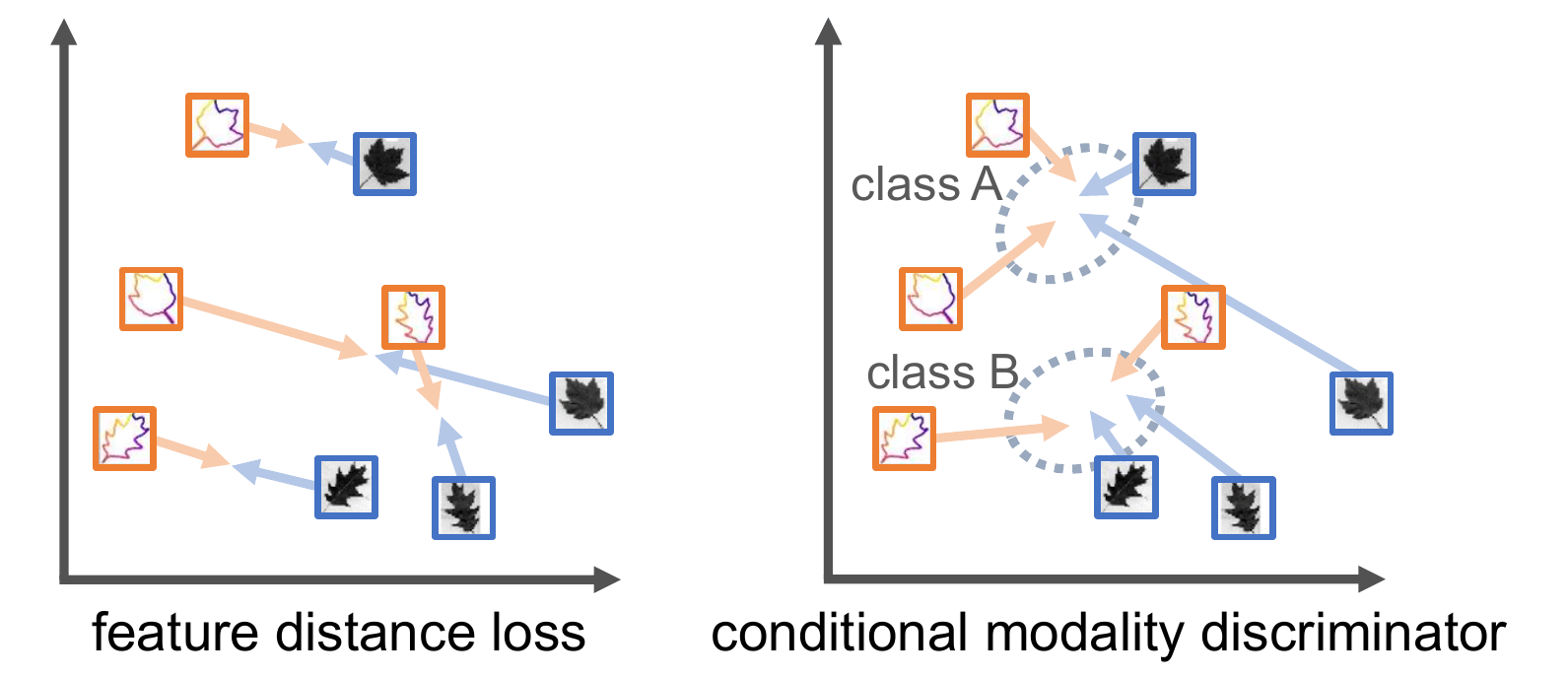}\\[-3mm]
    \caption{The role of the feature distance loss and the conditional modality discriminator. Blue indicates the original modality (image) and orange indicates the augmented modality (time series).}
    \label{fig:share}
\end{figure}

\subsection{Feature Distance Loss}
In order to directly correlate the features of each modality and create hard consistent features, we use a feature distance loss. 
The objective of this loss is to add a hard consistency in the embedding between pairs of patterns, as shown in Fig.~\ref{fig:share}.
The feature distance loss $\mathcal{L}_\mathrm{FD}$ is the Mean Squared Error (MSE) between original embedding $f_\mathrm{org}$ and self-augmented embedding $f_\mathrm{aug}$, i.e., 
\begin{equation}
\mathcal{L}_\mathrm{FD} =  ||f_\mathrm{org} - f_\mathrm{aug}||^2/2,\end{equation}
where $||\cdot||$ is the $L^2$ distance.
This loss contributes in optimizing the encoders to try to have the same encoding for the same pattern across both modalities.

\subsection{Conditional Modality Discriminator}\vspace{-2mm}
To make $f_\mathrm{org}$ and $f_\mathrm{aug}$ softly consistent, so that they keep the characteristics of their complementary modalities,  
we use a conditional modality discriminator (CMD). The CMD is based on a conditional Generative Neural Net (cGAN)~\cite{mirza2014conditional} and inspired by cross-modal GANs with modality discriminators~\cite{Wang_2017adversarial,Peng_2019,x-gacmn}. 
By this framework, the two modalities become indistinguishable in $\mathcal{F}$ without requiring a hard consistency between $f_\mathrm{org}$ and $f_\mathrm{aug}$.  

Especially for the application to the classification tasks, 
CMD can be class-conditional, where the class label of the pattern $x_\mathrm{org}$ is provided. 
This is done by concatenating a one-hot vector of the class to the input of the CMD.
As shown in Fig.~\ref{fig:share}, the condition encourages the embeddings to not only be modality invariant but also class-consistent.\par

The encoders and the CMD play an adversarial min-max game. 
The CMD seeks to maximize the probability of discriminating one modality from the other. The loss for the discriminator $D$ (the CMD) is defined as:
\begin{equation}
\label{eq:cmd}
\mathcal{L}^D_\mathrm{CMD} = -{d\ \mathrm{log}(\hat{d})+(1-d)\mathrm{log}(1-\hat{d})},
\end{equation}
where $d\in \{0,1\}$ is the target label (target modality) and $\hat{d}\in [0,1]$ is the predicted label by the discriminator. 
During training, the target modality alternates between the original and the self-augmented modalities. 
The encoders $E$ have a similar role as the generator in a GAN in that they try to minimize the log of the inverse probability predicted by discriminator $D$ and its loss is defined as:  
\begin{equation}
\label{eq:cmd2}
\mathcal{L}^E_\mathrm{CMD} = -\mathrm{log}(1-\hat{d}).
\end{equation}
By this GAN-based mechanism, the encoders can create an encoding in which the original modality is indistinguishable from the self-augmented modality, but not necessarily require that the original pattern and the self-augmented pattern be embedded near each other in the space.

\section{Classification Using a Gating Neural Network}
\label{sec:gating} In order to demonstrate the effectiveness of the proposed embedding and to perform robust classification, the shared feature space embeddings, $f_\mathrm{org}$ and $f_\mathrm{aug}$, are combined using a gating network and classified using a neural network, as shown in Fig.~\ref{fig:overall}. \par

Our gating mechanism controls the flow of information from each modality and combines them in a simple weighted sum, or:
\begin{equation}
\label{eq:alpha}
f=\alpha f_\mathrm{org} + (1-\alpha)f_\mathrm{aug},
\end{equation}
where $f$ is the combined embedding and $\alpha$ is a trained weighting parameter.
The value of $\alpha$ is trained simultaneously with the encoders and classifier using a separate gating network. 
It should be emphasized that our embedding scheme allows us to use this simple gating mechanism --- unlike typical multi-modal fusion methods, such as concatenation, the proposed shared space embedding creates the element-wise correspondence between the features $f_\mathrm{org}$ and $f_\mathrm{aug}$ and thus their direct addition is possible. In other words, independently trained networks will not have corresponding features and might not be able to be combined in this way.\par
The classification by $f$ is a Multi-Layer Perceptron (MLP), which is, in practice, realized as two fully-connected layers connected to the two encoders via the above weighted sum operation. 
The interaction between the encoders and classifier is similar to a Convolutional Neural Network (CNN).
Note that the encoders, gating network, and classifier are trained in one training step and the CMD is trained in an alternate training step.\par

\section{Experimental Results}
\label{sec:results}
\subsection{Datasets}\vspace{-2mm}
In order to evaluate the proposed method, we use two types of datasets with different modality: online handwriting datasets and a leaf image dataset. The online handwriting patterns are time series and their self-augmented counterparts are images. 
In contrast, the leaf images are image and their self-augmented counterparts are time series. By using those datasets, we can observe the performance of the proposed method in both directions of Fig.~\ref{fig:concept}.\par

The online handwriting dataset used for the experiments is the Unipen multi-writer isolated digit (1a), uppercase character (1b), and lowercase character (1c) datasets\footnote{\url{http://www.unipen.org/}}. 
The datasets were divided into 80$\%$ for training and 20$\%$ for testing.  
For the self-augmentation, a binary $32\times32$ image was rendered from each time-series trajectory.
\par
The leaf dataset, OSULeaf\footnote{\url{http://web.engr.oregonstate.edu/~tgd/leaves/}}, is made of images of 6 types of leaves and has a pre-set training and test set of 200 and 242 patterns, respectively. 
For this dataset, pseudo-time series are extracted from the contours. 
They are represented by 2D coordinates sampled to 50 time steps. 
In addition, the training set was augmented by 60 times via random rotations in increments of 6 degrees. 

\subsection{Network Structure and Training Protocol}\vspace{-2mm}
The proposed method model is composed of five neural networks: two encoders, the CMD, a gating network, and a classifier. 
The encoders have three convolutional layers with kernel size 3, batch normalization, Rectified Linear Unit (ReLU) activations, and max pooling. They have 32, 64, and 128 filters each.
The time series-based encoder uses 1D convolutions and the image-based encoder uses 2D convolutions. 
The encoders also have two fully connected layers with 512 nodes, batch normalization, and ReLU activations.
The gating network is identical to the decoders, except that the output of the last convolutional layer is fused using concatenation.
The classifier is an MLP with 512 nodes, batch norm, and ReLU.
The CMD is similar to the classifier except with an extra fully-connected layer of 512 nodes plus the one-hot representation of the classes.\par
%

The networks are trained using iterations with two steps. These two steps are based on the training protocol of GAN. In the first step of the iteration, the CMD is trained using Eq.~\eqref{eq:cmd} and then the weights of the encoders are fixed. This step is the same as the step of learning the discriminator of the general GAN. In the second step, the weights of the CMD are fixed, and the encoders, gating network, and classifier are trained using Eq.~\eqref{eq:cmd2}, the feature distance loss, and classification loss. This step is the same as the step of learning the generator of the general GAN, but in the proposed method, it is different in that the gating network and classifier are trained simultaneously using feature distance loss and classification loss. For training, Adam optimizer is used with an initial learning rate of 0.0001 for 400 epochs.

\begin{table}[t]
\caption{Classification Accuracy (\%)}
\label{table:result}
\begin{center}
\begin{small}
\begin{tabular}{lccccc}
 \multicolumn{3}{l}{\textbf{Method}\hspace{1cm}  \textbf{Unipen 1a}}    & \textbf{1b}    & \textbf{1c}    & \textbf{OSULeaf} \\ \hline
\multicolumn{2}{l}{Proposed} & 98.75          & 98.51          & 96.96 & 97.93            \\
\multicolumn{2}{l}{\quad w/o CMD} & 98.62          & 98.57 & 96.87          & 96.69            \\
\multicolumn{2}{l}{\quad w/o $\mathcal{L}_\mathrm{FD}$} & 98.79          & 97.84          & 96.16          & 97.52            \\
\multicolumn{2}{l}{CNN (image)}             & 98.14          & 97.51          & 94.58          & 94.63            \\
\multicolumn{2}{l}{CNN (time series)}       & 98.23          & 96.90          & 96.26          & 92.98            \\
\multicolumn{2}{l}{CNN (concat)}        & 98.66          & 98.24          & 96.84          & 95.87           \\
\hline
\multicolumn{2}{l}{LDF-Span.~\cite{Kenji_Iwana_2020}} & 98.87 & 98.07 & 97.20 & 65.3 \\
\multicolumn{2}{l}{Google~\cite{Keysers_2017}} & 99.2 & 96.9 & 94.9 & -- \\
\multicolumn{2}{l}{CNN DWA~\cite{Iwana_2019}} & 98.5 & 96.1 & 95.9 & -- \\
\multicolumn{2}{l}{LSTM~\cite{Iwana_2019}} & 96.8 & 92.3 & 89.8 & -- \\
\multicolumn{2}{l}{ResNet~\cite{Wang_2017}} & -- & -- & -- & 97.9 \\
\multicolumn{2}{l}{LSTM-FCN~\cite{Karim_2018}} & -- & -- & -- & 99.59 \\
\multicolumn{2}{l}{G-G-MTF~\cite{wang2015imaging}} & -- & -- & -- & 64.2 \\
\hline
\end{tabular}
\end{small}
\end{center}
\vspace{-4mm}
\end{table}

\subsection{Quantitative Evaluation}\vspace{-2mm}
The classification performance of the proposed method is compared with state-of-the-art time series classification models. We also compare our method with two single-modality CNNs, called CNN (image) and CNN (time series), and their direct feature concatenation model, CNN (concat). Those CNNs have the same structure as the encoder in the proposed method. As ablation studies, we also evaluate the proposed method trained without the CMD or $\mathcal{L}_\mathrm{FD}$. \par
Table \ref{table:result} shows the results. For all datasets, the proposed method performed better than or comparative to all the comparative methods. More importantly, we can confirm the following three facts from this result:
\begin{itemize}\setlength{\itemsep}{0pt}
\item Usefulness of the self-augmentation is confirmed by comparing the proposed method with CNN (time series) for three Unipen datasets and CNN (image) for OSULeaf. Especially for OSULeaf, we could reduce about 61\% of the misrecognitions by using the self-augmented modality.
\item Usefulness of the shared feature space embedding is also established by comparing the proposed method with CNN (concat). 
\item Use of both hard and soft consistencies gives slight accuracy improvements. Although both consistencies try to make $f_\mathrm{org}\sim f_\mathrm{aug}$ and thus have similar effects on the feature embedding, the difference in their strength still realizes their complementary role.   
\end{itemize} 

\subsection{Observation of Feature Distributions}
\begin{figure}[t]
    \centering
     \includegraphics[width=0.48\textwidth]{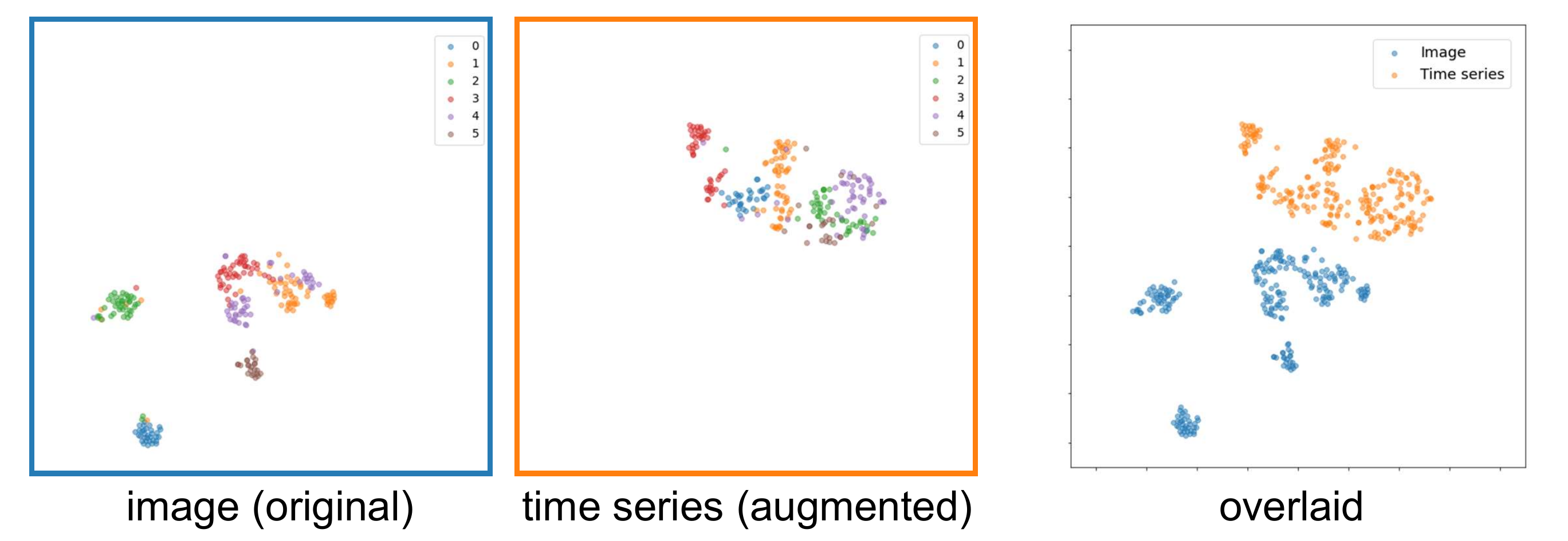}\\
    {\small (a)~Independent feature embeddings of two modalities}\\ \smallskip
\includegraphics[width=0.48\textwidth]{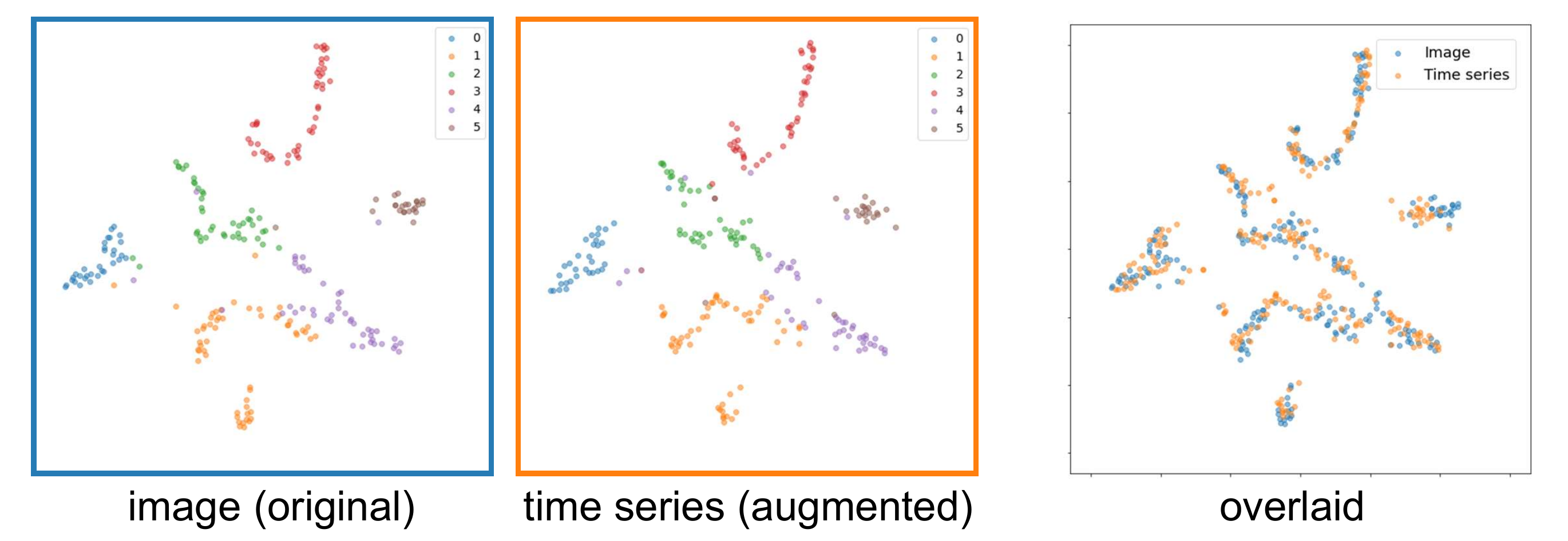}\\
    {\small (b)~Shared feature space embedding}\\[-2mm]
    \caption{Feature distributions by t-SNE for OSULeaf.}
    \label{fig:tSNE}
\end{figure}

Fig.~\ref{fig:tSNE} visualizes the feature distributions of two modalities, i.e., the original image modality and the self-augmented time-series modality, of OSULeaf. Without the consistency control between two modalities, their distributions become totally independent.  
In contrast, with the hard and the soft consistency controls, the distribution of $f_\mathrm{org}$ and $f_\mathrm{aug}$ are very similar as expected (but not perfectly identical, as we will see in \ref{sec:gate-effect}). 

\subsection{Observation of $\alpha$ Inferred by the Gating Network\label{sec:gate-effect}}

\begin{figure}[t]
    \centering
    \includegraphics[width=1.0\linewidth]{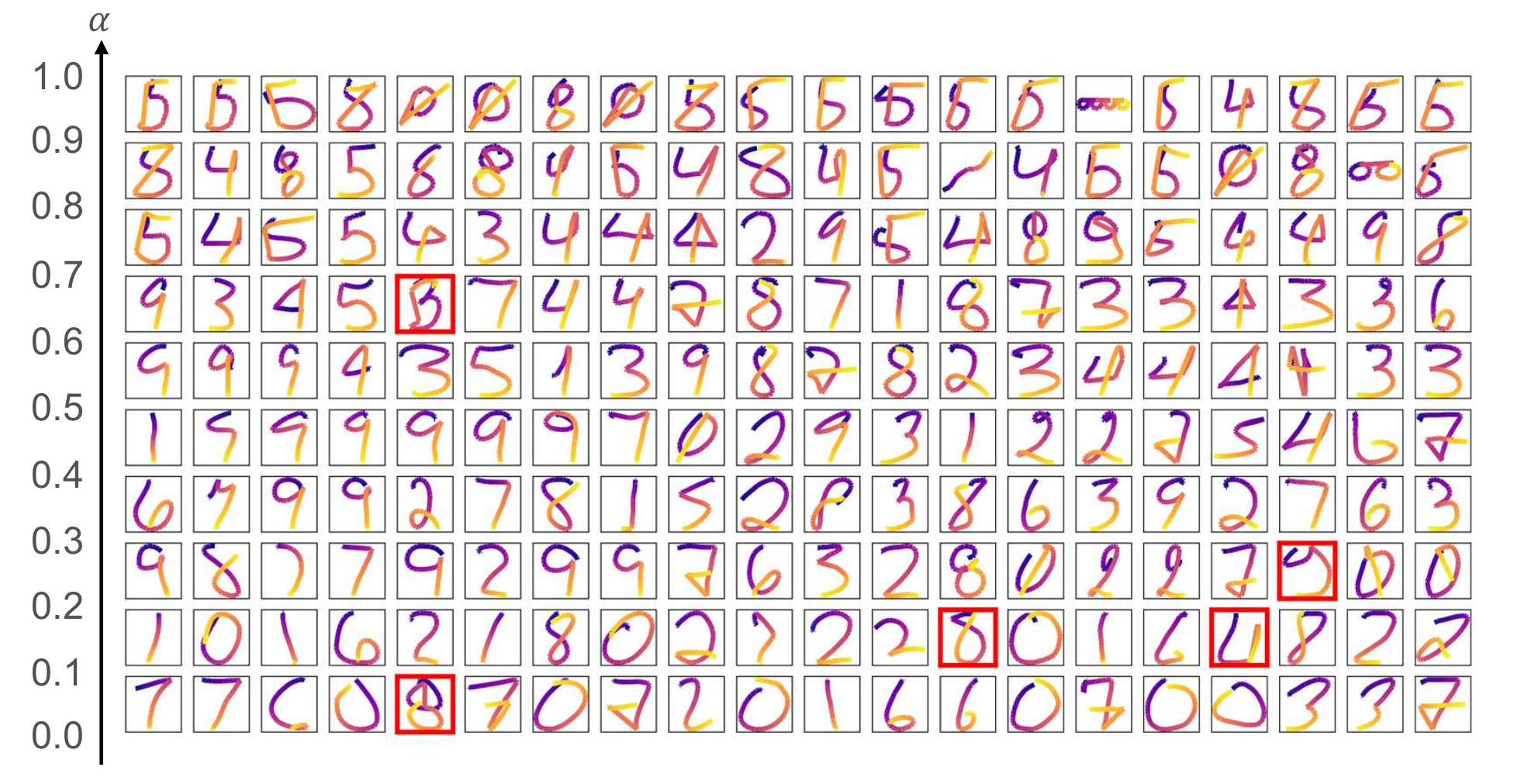}\\[-3mm]
    \caption{Unipen 1a test patterns that give $\alpha\in\{0,0.1,\ldots,0.9, 1.0\}$. 20 patterns are randomly selected for each $\alpha$. Misclassified patterns are in red box.}
    \label{fig:alpha}
\end{figure}

This section shows that the feature embeddings $f_\mathrm{org}$ and $f_\mathrm{aug}$ are not perfectly identical and 
keep the characteristics of their modalities. In fact, if $f_\mathrm{org}=f_\mathrm{aug}$ by the hard consistency 
by the feature distance loss, the gating mechanism (Eq. \eqref{eq:alpha}) does not make any sense. However, 
Fig.~\ref{fig:alpha}, which shows the relationship between patterns and their $\alpha$, experimentally proves that $f_\mathrm{org}$ and $f_\mathrm{aug}$ are not perfectly identical. Specifically, when $\alpha=0$, only $f_\mathrm{aug}$ (self-augmented image modality feature) is used. It is reasonable that class pairs (e.g., ``0''-``6'' and ``1''-``7'') which are confusing in the time-series modality tend to take $\alpha\sim 0$. It is also reasonable that class pairs (e.g., ``5''-``8'' and ``(slashed) 0'' and ``8'') which are confusing in the image modality tend to take larger $\alpha\sim 1$. This fact reveals that the proposed method utilizes the characteristics of two modalities appropriately while taking the advantage of the modality consistencies.

\subsection{Improved and Deteriorated Samples}

\begin{figure}
    \centering
    \includegraphics[width=0.30\textwidth]{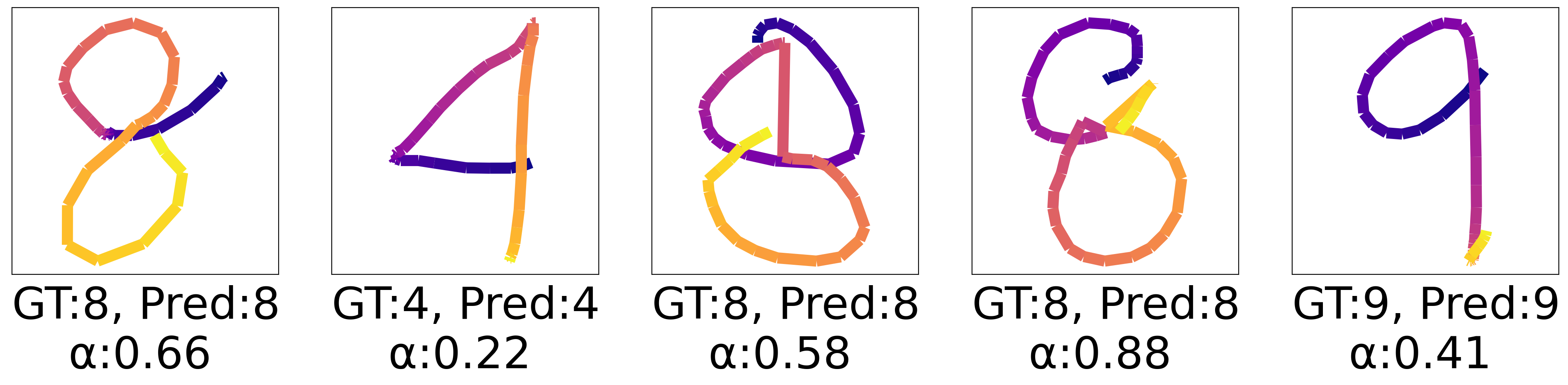}\\ \vspace{-1mm}
    {\small (a)~Patterns with an uncommon writing order}\\ \smallskip \vspace{1mm}
    \includegraphics[width=0.24\textwidth]{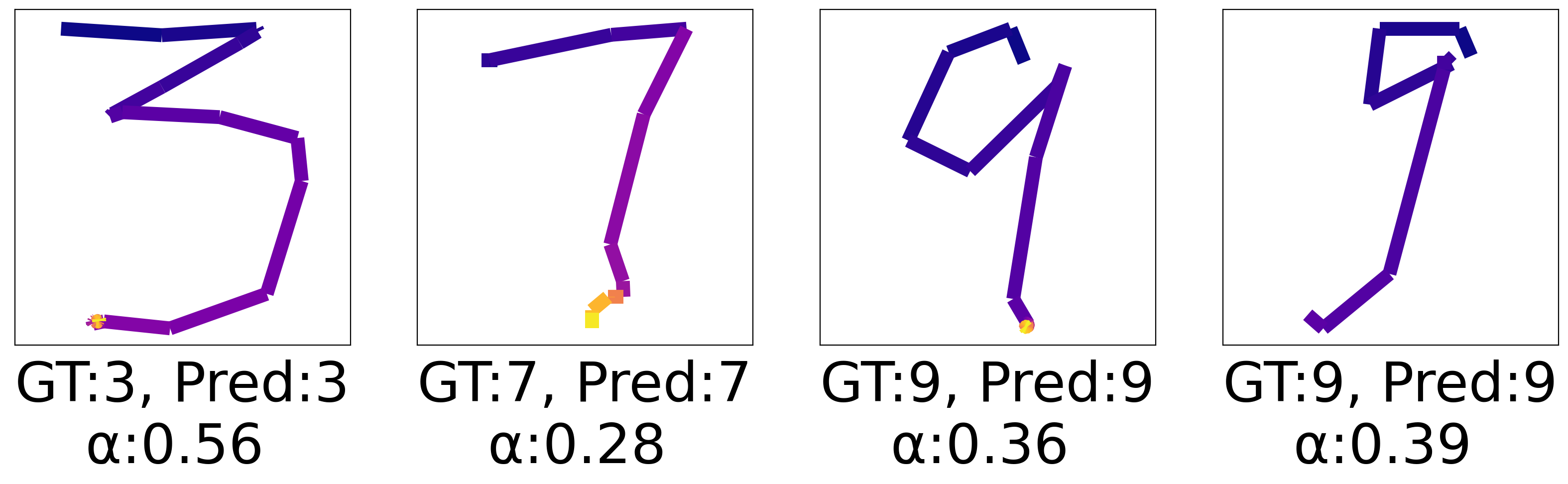}\\ \vspace{-1mm}
    {\small (b)~Patterns written quickly for most of the pattern} \vspace{-1mm}
    \caption{Improved samples where the proposed method correctly predicted the pattern and where CNN (time series) misclassified them. ``GT'' is the ground truth. ``Pred'' is the prediction of the proposed method.}
    \label{fig:improved_samples_ts}
\end{figure}

\begin{figure}
    \begin{minipage}{0.68\hsize}
    \begin{center}
    \includegraphics[width=0.95\textwidth]{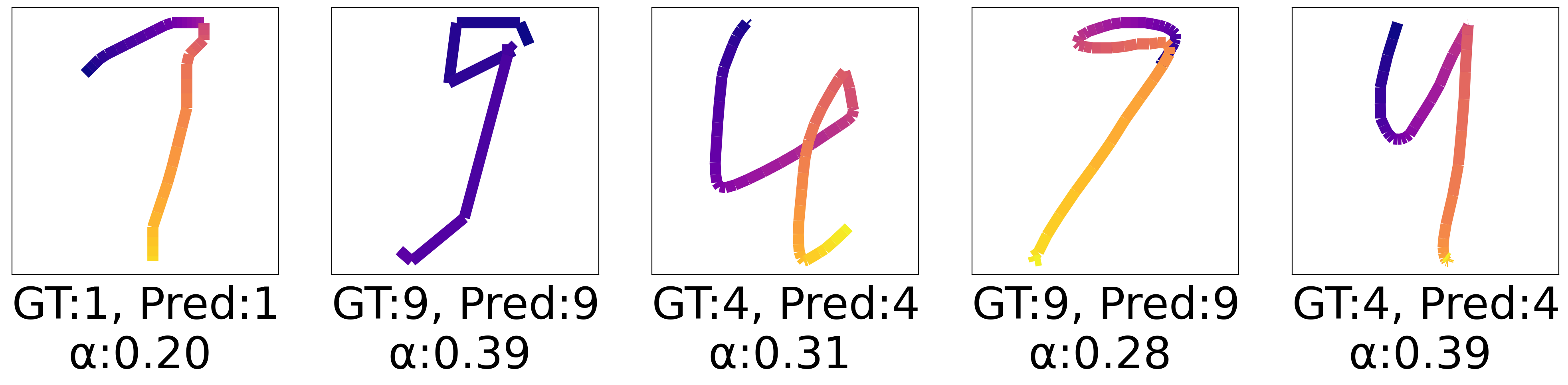}
    {\small (a)~Small $\alpha$}
    \end{center}
    \end{minipage}
    \begin{minipage}{0.31\hsize}
    \begin{center}
    \includegraphics[width=0.8\textwidth]{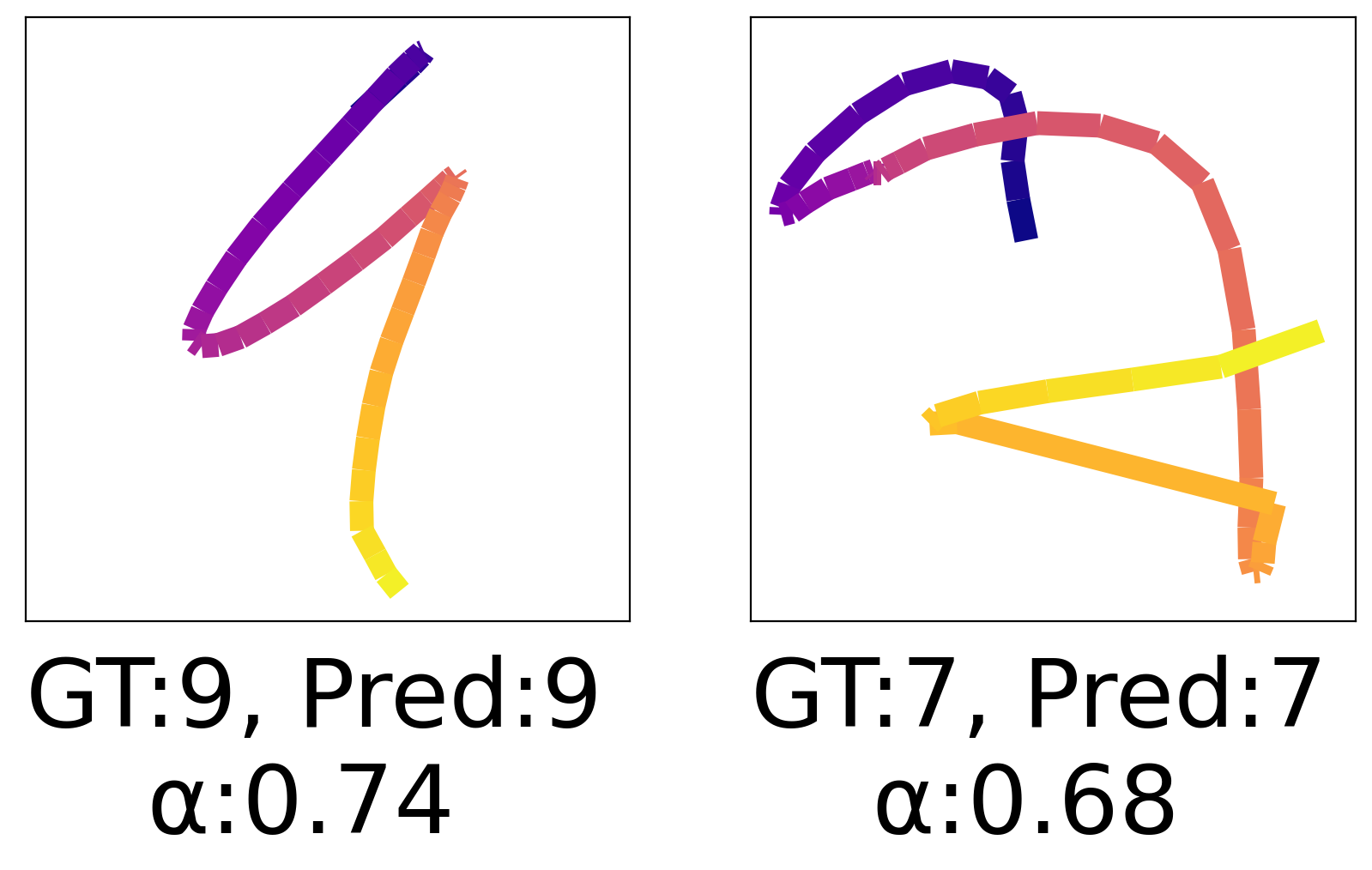}
    {\small (b)~Large $\alpha$}
    \end{center}
    \end{minipage} 
    \vspace{-3mm}
    \caption{Improved samples where the proposed method correctly predicted the pattern and where CNN (concat) misclassified them. ``GT'' is the ground truth. ``Pred'' is the prediction of the proposed method.}
    \label{fig:improved_samples_concat}
\end{figure}

\begin{figure}
    \centering
    \includegraphics[width=0.24\textwidth]{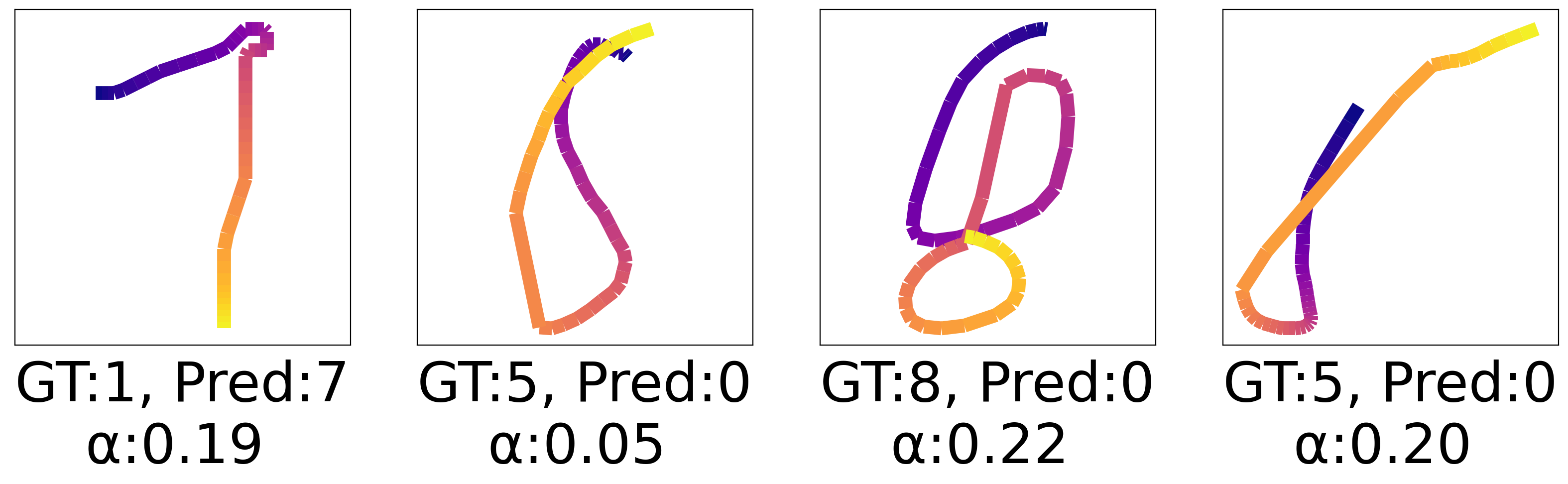}
    \vspace{-3mm}
    \caption{Deteriorated samples where the proposed method incorrectly predicted the pattern. ``GT'' is the ground truth. ``Pred'' is the prediction of the proposed method.}
    \label{fig:deteriorated_samples}
\end{figure}

Improved samples for Unipen 1a (digit) are shown in Figs.~\ref{fig:improved_samples_ts},~\ref{fig:improved_samples_concat}. The recognition results of the proposed method were compared with those of CNN (time series) and CNN (concat).

Improved samples between the proposed method and CNN (time series) show that the proposed method can use the self-augmented image modality effectively. There were two cases in the improved samples. One is patterns with an uncommon writing order, and the other is patterns written quickly for most of pattern. The results are shown in Fig.~\ref{fig:improved_samples_ts} (a) and (b), respectively. Both of them are difficult to classify from the viewpoint of time series data because they are very different from typical patterns. On the other hand, from the viewpoint of image data, they are not so different. Therefore, the proposed method is more robust to temporal distortions than CNN (time series) because of the use of self-augmented image modality.

Improved samples between the proposed model and CNN (concat) show that the proposed method can control the features of each modality (original and self-augmented modality). Fig.~\ref{fig:improved_samples_concat} (a) shows improved samples with a small alpha, that is, using more features of the self-augmented image modality. This also means that the proposed method used fewer features of the original time series modality than CNN (concat).  Fig.~\ref{fig:improved_samples_concat} (b) shows the opposite.

However, there were deteriorated patterns due to the wrong control of modality features. This is shown in Fig.~\ref{fig:deteriorated_samples}. In these cases, $\alpha$ is small, meaning that they rely on the image modality. 
In these cases, due to their reliance on the augmented image modality, they were misclassified.

\section{Conclusion}
\label{sec:conc}
In this paper, we proposed a method of self-augmented multi-modal feature embedding, that enables the utilization of complementary characteristics of different modalities under a unified feature representation. The usefulness of the extracted feature is confirmed quantitatively and qualitatively via classification experiments with not only image patterns (whose self-augmented modality is time series) but also time series patterns (whose self-augmented modality is an image). Since self-augmentation is a very general idea and requires no extra cost for the data acquisition, it is possible to use it for enhancing the accuracy of classification tasks and 
other signal analysis tasks. 

\newpage







\bibliographystyle{IEEEbib}
\bibliography{main}

\begin{thebibliography}{10}

\bibitem{duong2017multimodal}
Chi~Thang Duong, Remi Lebret, and Karl Aberer,
\newblock ``Multimodal classification for analysing social media,''
\newblock {\em arXiv preprint arXiv:1708.02099}, 2017.

\bibitem{ortega2019multimodal}
Juan~DS Ortega, Mohammed Senoussaoui, Eric Granger, Marco Pedersoli, Patrick
  Cardinal, and Alessandro~L Koerich,
\newblock ``Multimodal fusion with deep neural networks for audio-video emotion
  recognition,''
\newblock {\em arXiv preprint arXiv:1907.03196}, 2019.

\bibitem{lucieri2020benchmarking}
Adriano Lucieri, Huzaifa Sabir, Shoaib Ahmed~Siddiqui, Syed~Tahseen Raza~Rizvi,
  Brian~Kenji Iwana, Seiichi Uchida, Andreas Dengel, and Sheraz Ahmed,
\newblock ``Benchmarking deep learning models for classification of book
  covers,''
\newblock {\em SN Comp. Sci.}, vol. 1, no. 139, pp. 1--16, 2020.

\bibitem{arevalo2017gated}
John Arevalo, Thamar Solorio, Manuel Montes-y G{\'o}mez, and Fabio~A
  Gonz{\'a}lez,
\newblock ``Gated multimodal units for information fusion,''
\newblock {\em arXiv preprint arXiv:1702.01992}, 2017.

\bibitem{park2016image}
Gwangbeen Park and Woobin Im,
\newblock ``Image-text multi-modal representation learning by adversarial
  backpropagation,''
\newblock {\em arXiv preprint arXiv:1612.08354}, 2016.

\bibitem{Wei_2020}
Kaimin Wei and Zhibo Zhou,
\newblock ``Adversarial attentive multi-modal embedding learning for image-text
  matching,''
\newblock {\em {IEEE} Access}, vol. 8, pp. 96237--96248, 2020.

\bibitem{Wang_2017adversarial}
Bokun Wang, Yang Yang, Xing Xu, Alan Hanjalic, and Heng~Tao Shen,
\newblock ``Adversarial cross-modal retrieval,''
\newblock in {\em ACM MM}, 2017.

\bibitem{Wimmer2008}
Matthias Wimmer, Bj{\"o}rn Schuller, Dejan Arsic, Bernd Radig, and Gerhard
  Rigoll,
\newblock ``Low-level fusion of audio and video feature for multi-modal emotion
  recognition,''
\newblock in {\em VISAPP}, 2008.

\bibitem{brady2016multi}
Kevin Brady, Youngjune Gwon, Pooya Khorrami, Elizabeth Godoy, William Campbell,
  Charlie Dagli, and Thomas~S Huang,
\newblock ``Multi-modal audio, video and physiological sensor learning for
  continuous emotion prediction,''
\newblock in {\em IWAVEC}, 2016, pp. 97--104.

\bibitem{Peng_2019}
Yuxin Peng and Jinwei Qi,
\newblock ``{CM}-{GANs}: Cross-modal generative adversarial networks for common
  representation learning,''
\newblock {\em {ACM} Trans. Multimedia Comput., Com., and Appl.}, vol. 15, no.
  1, pp. 1--24, 2019.

\bibitem{Wang_2019}
Hao Wang, Doyen Sahoo, Chenghao Liu, Ee~peng Lim, and Steven C.~H. Hoi,
\newblock ``Learning cross-modal embeddings with adversarial networks for
  cooking recipes and food images,''
\newblock in {\em CVPR}, 2019.

\bibitem{Huang_2018}
Feiran Huang, Xiaoming Zhang, Chaozhuo Li, Zhoujun Li, Yueying He, and Zhonghua
  Zhao,
\newblock ``Multimodal network embedding via attention based multi-view
  variational autoencoder,''
\newblock in {\em ACM ICMR}, 2018.

\bibitem{Spurr_2018}
Adrian Spurr, Jie Song, Seonwook Park, and Otmar Hilliges,
\newblock ``Cross-modal deep variational hand pose estimation,''
\newblock in {\em CVPR}, 2018.

\bibitem{sumi2019modality}
Taichi Sumi, Brian~Kenji Iwana, Hideaki Hayashi, and Seiichi Uchida,
\newblock ``Modality conversion of handwritten patterns by cross variational
  autoencoders,''
\newblock in {\em ICDAR}, 2019, pp. 407--412.

\bibitem{x-gacmn}
Weikuo Guo, Jian Liang, Xiangwei Kong, Lingxiao Song, and Ran He,
\newblock ``X-gacmn: An x-shaped generative adversarial cross-modal network
  with hypersphere embedding,''
\newblock in {\em ACCV}, 2018.

\bibitem{mirza2014conditional}
Mehdi Mirza and Simon Osindero,
\newblock ``Conditional generative adversarial nets,''
\newblock {\em arXiv preprint arXiv:1411.1784}, 2014.

\bibitem{Kenji_Iwana_2020}
Brian~Kenji Iwana and Seiichi Uchida,
\newblock ``Time series classification using local distance-based features in
  multi-modal fusion networks,''
\newblock {\em Pattern Recognition}, vol. 97, pp. 107024, 2020.

\bibitem{Keysers_2017}
Daniel Keysers, Thomas Deselaers, Henry~A. Rowley, Li-Lun Wang, and Victor
  Carbune,
\newblock ``Multi-language online handwriting recognition,''
\newblock {\em {IEEE} Trans. Pattern Anal. and Mach. Intel.}, vol. 39, no. 6,
  pp. 1180--1194, 2017.

\bibitem{Iwana_2019}
Brian~Kenji Iwana and Seiichi Uchida,
\newblock ``Dynamic weight alignment for temporal convolutional neural
  networks,''
\newblock in {\em ICASSP}, 2019.

\bibitem{Wang_2017}
Zhiguang Wang, Weizhong Yan, and Tim Oates,
\newblock ``Time series classification from scratch with deep neural networks:
  A strong baseline,''
\newblock in {\em IJCNN}, 2017.

\bibitem{Karim_2018}
Fazle Karim, Somshubra Majumdar, Houshang Darabi, and Shun Chen,
\newblock ``{LSTM} fully convolutional networks for time series
  classification,''
\newblock {\em {IEEE} Access}, vol. 6, pp. 1662--1669, 2018.

\bibitem{wang2015imaging}
Zhiguang Wang and Tim Oates,
\newblock ``Imaging time-series to improve classification and imputation,''
\newblock in {\em IJCAI}, 2015, pp. 3939--3945.

\end{thebibliography}

\end{document}